\documentclass{article}

\usepackage{iclr2026_conference,times}

\usepackage[utf8]{inputenc} %
\usepackage[T1]{fontenc}    %
\usepackage{hyperref}       %
\usepackage{url}            %
\usepackage{booktabs}       %
\usepackage{amsfonts}       %
\usepackage{nicefrac}       %
\usepackage{microtype}      %
\usepackage{xcolor}         %
\usepackage{graphicx}
\usepackage{tikz}
\usepackage{subcaption}
\usepackage{wrapfig}
\usepackage{amssymb}
\usepackage{enumitem}
\usepackage{amsmath}

\usepackage{booktabs}
\usepackage{multirow}

\usepackage{graphicx}
\usepackage{subcaption}  

\definecolor{darkblue}{rgb}{0, 0, 0.5}
\hypersetup{colorlinks=true, citecolor=darkblue, linkcolor=darkblue, urlcolor=darkblue}

\providecommand{\todo}[1]{
    {\protect\color{red}{[TODO: #1]}}
}

\providecommand{\danqi}[1]{
    {\protect\color{orange}{[Danqi: #1]}}
}
\providecommand{\dan}[1]{
    {\protect\color{teal}{[Dan: #1]}}
}

\providecommand{\adithya}[1]{
    {\protect\color{brown}{[Adithya: #1]}}
}

\providecommand{\todon}{
    {\protect\color{red}{00.00}}
}

\providecommand{\danqi}[1]{}
\providecommand{\dan}[1]{}
\providecommand{\adithya}[1]{}
\providecommand{\todo}[1]{}
\providecommand{\todon}[1]{}

\usepackage{amsmath,amsfonts}

\makeatletter
\newcommand{\ve}{\@ifnextchar\bgroup{\velong}{{\bm{e}}}}
\newcommand{\velong}[1]{{\bm{#1}}}
\makeatother

\def\ve{{\mathbf{e}}}

\DeclareMathAlphabet{\mathsfit}{\encodingdefault}{\sfdefault}{m}{sl}
\SetMathAlphabet{\mathsfit}{bold}{\encodingdefault}{\sfdefault}{bx}{n}

\title{Extreme Adaptive Transformer for Time Series forecasting}

\author{%
    Sanjeev Shrestha \hspace{2em} Hui Liu \hspace{2em} Yifan Zhang \hspace{2em} \\
    Department of Computer Science, Missouri State University\\
    \texttt{\{ss472s, HuiLiu, YifanZhang\}@MissouriState.edu}
}

\iclrfinalcopy
\begin{document}

\maketitle

\begin{abstract}

Time series forecasting remains challenging when the underlying data contain rare but critical extreme events. This issue is particularly important in hydrologic forecasting, where streamflow distributions are often highly skewed and extreme peaks can have substantial impacts on flood monitoring, water resource management, and early warning systems. Although Transformer-based forecasting models have achieved strong performance by modeling long-range temporal dependencies, they typically treat all time points uniformly and may therefore underrepresent rare extreme patterns. In this paper, we propose the Extreme-Adaptive Transformer (Exformer), a forecasting framework designed to explicitly model temporal dependencies involving both normal and extreme events. Exformer introduces an extreme-adaptive attention mechanism composed of three sparse components: Local, Stride, and Extreme. The Local and Stride components capture short-term and periodic temporal dependencies, respectively, while the Extreme component selectively models event-aware dependencies between normal and extreme streamflow patterns. Experiments on four real-world hydrologic streamflow datasets show that Exformer achieves superior 3-day forecasting performance compared with state-of-the-art baselines. Our findings demonstrate that explicitly incorporating extreme-aware attention improves the forecasting capacity of Transformer models on imbalanced time series with rare but consequential events. Code is available at \href{https://github.com/sanzexstha/Exformer}{https://github.com/sanzexstha/Exformer}.
\end{abstract}

\section{Introduction}

Multivariate time series (MTS) forecasting plays a crucial role in many real-world domains, including hydrology~\citep{pfformer}, transportation~\citep{Traffic_deep}, finance~\citep{stock_pred}, and environmental monitoring~\citep{weather_deep}. 
In long-term time series forecasting, extreme events are infrequent but particularly important. For example, sudden streamflow peaks caused by heavy rainfall occur rarely but can substantially affect forecasting performance and real-world decision-making. Consequently, accurately predicting streamflow under extreme conditions is critical for water resource management, flood warning, drought monitoring, and public safety. However, forecasting extreme events remains challenging because extreme conditions provide substantially fewer observations than normal conditions.

Traditional statistical methods, such as autoregressive models, moving average models, exponential smoothing, and ARIMA-based variants, have been widely used for time series forecasting~\citep{box}. Although these methods can be effective for relatively stable series, they often struggle to model nonlinear temporal dependencies, high variance, and extreme values. More recently, deep learning models, including recurrent neural networks~\citep{rnn}, hybrid architectures~\citep{hybrid}, and graph neural networks~\citep{wugnn}, have been developed to improve forecasting performance by learning complex temporal representations from data. Among them, Transformer-based models have been widely studied for time series forecasting because of their ability to capture long-range dependencies~\citep{timeseriessurvey}. Nevertheless, general forecasting models can still perform poorly on highly skewed data, particularly when the target series contains rare extreme events that are underrepresented relative to normal observations.

A key issue in extreme-event forecasting is the imbalance between normal and extreme observations. Since most time points correspond to normal values, a forecasting model can obtain low average error while still failing to capture the most important peaks. This limitation is especially problematic in hydrologic forecasting, where extreme streamflow values may correspond to flood-related events. Previous extreme-adaptive approaches have attempted to address this issue by designing models that place greater emphasis on extreme values. DAN~\citep{dan} introduced a polar representation learning framework that separately models far and near representations and uses a distance-weighted multi-loss objective to improve robustness to extreme events. PFformer~\citep{pfformer} further improved multivariate streamflow forecasting by using position-free embedding strategies, including Enhanced Feature-based Embedding and Auto-Encoder-based Embedding, to better capture dependencies between streamflow and auxiliary variables such as rainfall.

Despite these advances, existing attention mechanisms are generally token-agnostic with respect to event severity and therefore do not condition query-key selection on whether a token corresponds to a normal or extreme event. Standard full attention computes interactions over all query-key pairs, resulting in quadratic complexity and potentially introducing many low-relevance interactions that dilute attention to rare extreme-event patterns. Sparse attention mechanisms, such as local or stride attention~\citep{dozer}, reduce computational cost by restricting the attention pattern, but they do not explicitly preserve or model dependencies among extreme-event tokens.

To address the aforementioned limitations, we propose Exformer, an Extreme-Adaptive Transformer for multivariate time series forecasting. Exformer introduces an extreme-adaptive attention mechanism that dynamically adjusts the candidate key set according to whether a query corresponds to a normal or extreme time step. For normal time steps, Exformer employs Local and Stride attention components~\citep{dozer} to capture short-range temporal patterns and periodic dependencies among normal observations. For extreme time steps, Exformer incorporates an Extreme Attention component that enables each extreme query to selectively attend to other extreme keys. By integrating these components, Exformer reduces redundant query-key computations while emphasizing the temporal patterns most relevant to extreme-event forecasting.

Our main contributions are summarized as follows:
\begin{itemize}
    \item We introduce an Extreme-Adaptive Attention mechanism composed of Local, Stride, and Extreme components. Unlike static sparse attention mechanisms, the proposed mechanism adaptively selects query-key interactions according to whether each query corresponds to a normal or extreme input token, enabling the model to capture both regular temporal dependencies and rare extreme-event patterns while reducing the computational cost.
    
    \item We propose Exformer, an encoder-only Transformer framework for long-term time series forecasting. By incorporating Extreme-Adaptive Attention, Exformer provides a lightweight yet effective architecture for modeling imbalanced time series data.

    \item We conduct experiments on four real-world hydrologic streamflow datasets and compare Exformer with recent state-of-the-art baselines. The results demonstrate that Exformer achieves the best results in most RMSE and MAPE comparisons while reducing attention computation compared with full-attention baselines. In addition, ablation studies verify the effectiveness of the proposed Extreme-Adaptive Attention mechanism.

    % \item We evaluated Exformer on four real-world hydrologic time series datasets and showed that it outperforms state-of-the-art baselines, including DAN and PFformer, in forecasting accuracy while requiring substantially fewer FLOPs than several Transformer- and LSTM-based models.
\end{itemize}

    % \item We incorporate extreme-event labels derived from a Gaussian Mixture Model and use Kruskal-Wallis sampling to address the imbalance between normal and extreme training samples.

\section{Related Work}
\label{sec:related_work}

\subsection{Time Series Forecasting}
Time series forecasting has been extensively studied using both statistical and deep learning methods. Classical approaches, including autoregressive models, moving average models, exponential smoothing, VAR, and ARIMA-based variants, provide simple and interpretable forecasting frameworks. However, these methods typically rely on linear assumptions and often have limited capacity to model nonlinear temporal dependencies, complex multivariate interactions, and highly skewed distributions. In hydrologic forecasting, hybrid methods that combine decomposition techniques with ARIMA-based models have also been explored for long-term streamflow prediction, but they are not specifically designed to capture rare yet critical extreme values.

Deep learning methods have become increasingly popular for time series forecasting because of their ability to learn nonlinear temporal representations from data. Recurrent neural networks have been used to model sequential dependencies, while convolutional and graph-based models have been applied to capture local temporal patterns and cross-variable relationships. More recently, Transformer-based models have been widely adopted for time series forecasting due to their ability to capture long-range dependencies through self-attention. Despite their promising performance, general forecasting models are often optimized for overall prediction accuracy and may underrepresent sparse but important peaks in datasets with rare and severe extreme events. This limitation motivates the development of forecasting models that explicitly account for extreme-event distributions and event-aware temporal dependencies.

% Deep learning methods have become increasingly popular for time series forecasting because of their ability to model nonlinear patterns. Recurrent neural networks and LSTM-based models have been used to capture temporal dependencies, while convolutional and graph-based models have been applied to extract local and cross-variable patterns. Transformer-based models have also been widely adopted for long-term forecasting. Informer~\citep{informer} improves efficiency through ProbSparse self-attention and a generative decoder. Autoformer uses decomposition and autocorrelation mechanisms to model long-term dependencies. FEDformer~\citep{fedformer} represents time series in the frequency domain to preserve global patterns. More recent models such as PatchTST~\citep{patchTST}, Crossformer~\citep{crossformer}, and iTransformer~\citep{itransformer} further improve time series forecasting by introducting patching, cross-variable dependency modeling, and variate-token representations.

\subsection{Extreme-Adaptive Time Series Forecasting}

Extreme-event forecasting is challenging because extreme values are rare, imbalanced, and often exhibit patterns that differ from those of normal observations. In hydrologic time series, this problem is amplified by high skewness and kurtosis, with most values near normal flow levels, but a small number of values corresponding to severe streamflow peaks. Several studies have explored extreme-aware learning to improve forecasting in such conditions.

% The eGRU~\citep{egru} model extends the vanilla GRU by maintaining separate hidden states for normal and extreme events, enabling the model to capture distinct temporal patterns for each type of observation. Similarly, NEC+~\citep{nec} models the distributions of normal and extreme events and trains three predictors in parallel to improve robustness to extreme hydrologic events. DAN~\citep{dan} further extends extreme hydrologic forecasting by learning and merging rich polar representations. It is an LSTM-based model that learns polar representations by separating the far and near hidden spaces, thereby preserving information about extreme and normal values separately. It also uses a distance-weighted multi-loss objective, gate control vectors, and Kruskal-Wallis sampling to improve learning from imbalanced extreme data. This design demonstrates that explicitly modeling extreme patterns can improve long-horizon hydrologic forecasting.

eGRU~\citep{egru} extends the vanilla GRU by using separate hidden states for normal and extreme events, allowing distinct temporal patterns to be captured. NEC+~\citep{nec} models the distributions of normal and extreme events and trains multiple predictors to improve robustness to extreme hydrologic conditions. DAN~\citep{dan} further extends extreme hydrologic forecasting by learning polar representations that separate far and near hidden spaces, preserving information about extreme and normal values while using distance-weighted losses, gate control vectors, and Kruskal-Wallis sampling to handle data imbalance.

PFformer~\citep{pfformer} extends extreme-aware forecasting with a Transformer-based architecture. It introduces enhanced feature-based and auto-encoder-based embeddings to better model dependencies between streamflow and auxiliary variables such as rainfall. It also uses clustering-based oversampling and a multi-objective loss to improve performance during severe events.

The Exformer differs from DAN and PFformer by that it directly focuses on the attention mechanism. Instead of relying mainly on polar representation learning or position-free embeddings, Exformer modifies the self-attention pattern based on whether the query token is normal or extreme. This allows the model to preserve extreme-to-extreme dependencies while still efficiently modeling local and periodic dependencies among normal tokens.

\subsection{Sparse Attention Mechanisms}

Transformers have achieved remarkable performance in several domains, including natural language processing~\citep{transformer}, computer vision~\citep{image}, and time series analysis~\citep{timeseriessurvey}. However, the standard self-attention mechanism computes dot products between all query-key pairs, resulting in quadratic computational complexity in the input sequence length. This limits its scalability for long sequences and introduces redundant computation. To address this issue, several sparse attention mechanisms have been proposed. Longformer~\citep{longformer} restricts attention to local or dilated windows while allowing selected global tokens to attend across the sequence. BigBird~\citep{bigbird} combines local, random, and global attention patterns to reduce the number of query-key interactions. Informer~\citep{informer} introduces ProbSparse attention to focus on the most informative queries. Autoformer~\citep{autoformer} introduces a decomposition-based architecture and an Auto-Correlation mechanism to capture period-based temporal dependencies for long-term forecasting. FEDformer~\citep{fedformer} models time series in the frequency domain using Fourier- and Wavelet-based components, allowing it to capture global temporal patterns efficiently. Other time-series Transformer models improve efficiency through alternative mechanisms, such as patch-based representations in PatchTST~\citep{patchTST}, cross-variable modeling in Crossformer~\citep{crossformer}, and variate-token representations in iTransformer~\citep{itransformer}.

Dozer self-attention~\citep{dozer} is a sparse attention mechanism that captures temporal dependencies through local and stride-based attention patterns. Local attention focuses on nearby time steps, while stride attention captures periodic or seasonal dependencies. However, these sparse patterns do not explicitly distinguish between normal and extreme time steps. In datasets with rare extreme events, this can be limiting because extreme observations may contain important information that should be preserved even when they are temporally distant.

Extreme-Adaptive Attention builds on sparse temporal attention by introducing content-aware components. Its Local and Stride components capture regular temporal patterns, while its Extreme component allows queries corresponding to extreme tokens to selectively attend to keys associated with other extreme tokens. Unlike other static sparse attention mechanisms, the resulting attention structure can adapt to each input sequence, enabling the model to preserve rare but informative extreme-to-extreme dependencies while also reducing redundant computation.

\section{Method}

To address the aforementioned limitations, we propose Exformer, a forecasting framework equipped with an extreme-adaptive attention mechanism. In this section, we first introduce the encoder-only architecture of Exformer and present a step-by-step illustration of its input-to-output forecasting process. We then describe the proposed extreme-adaptive attention mechanism, which constructs sparse query-key interactions by preserving local, periodic, and extreme-event dependencies while filtering less informative attention pairs to improve forecasting accuracy and computational efficiency.

\subsection{Framework}

\begin{figure}[htbp]
\centering
   \includegraphics[width=\textwidth]{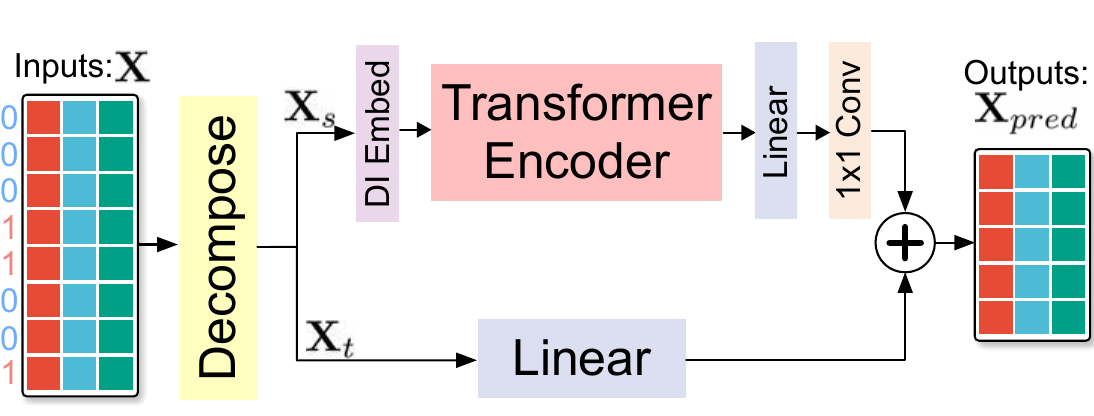}
\caption{
The architecture of our proposed Exformer framework.
}
\label{fig:Architecture}
\end{figure}

Given historical multivariate observations ($\mathbf{X} \in \mathbb{R}^{I \times D}$) with $I$ input time steps and $D$ variables, the MTS forecasting task aims to predict future values ($X_{pred} \in \mathbb{R}^{O \times D}$) over the next $O$ time steps. Figure~\ref{fig:Architecture} illustrates the overall framework of Exformer. Following previous decomposition-based forecasting methods~\citep{micn, autoformer, fedformer}, Exformer first decomposes the input sequence into seasonal and trend components, denoted as ($\mathbf{X}_s \in \mathbb{R}^{I \times D}$) and ($\mathbf{X}_t \in \mathbb{R}^{I \times D}$), respectively. The seasonal component is modeled by the Exformer encoder, while the trend component is forecast using a linear prediction layer.

The dimension-invariant embedding layer~\citep{multiscale} transforms the input MTS sequence into multi-channel feature maps while preserving the temporal and variable dimensions. It then partitions the sequence along the temporal dimension into non-overlapping patches, producing patched embeddings $\mathbf{X}_{enc} \in \mathbb{R}^{c \times N_{enc} \times p \times D}$, where $c$ denotes the number of embedded feature-map channels, $N_{enc} = \lceil I/p \rceil$ is the number of encoder patches, and $p$ is the patch size.
Since the proposed extreme-adaptive attention operates at the patch-token level, we assign a binary normal/extreme label to each patch. Specifically, an anomaly detection method first computes an outlier score $s_t$ for each time step. A threshold $\tau$ is then used to obtain a time-step-level label $\ell_t \in {0,1}$, where ($\ell_t=0$) denotes a normal time step and $\ell_t=1$ denotes an extreme time step. For each temporal patch $P_m$, the patch-level label $e_m$ is obtained by aggregating the time-step labels within the patch. The resulting patch labels are used to construct an extreme-aware mask that guides the attention mechanism to distinguish normal and extreme patch tokens.

The patched embeddings $\mathbf{X}_{enc} \in \mathbb{R}^{c \times N_{enc} \times p \times D}$, together with their patch-level normal/extreme labels, are fed into the Exformer encoder. The patch labels are used to construct the extreme-aware attention mask, enabling the encoder to model patch-level temporal dependencies with the proposed extreme-adaptive attention mechanism instead of canonical full attention. The encoder outputs from all patches are concatenated along the temporal dimension to reconstruct the full sequence representation, which is then projected from the input length to the prediction length using a linear layer. A $1 \times 1$ convolutional layer is applied to the learned latent representations to generate predictions for the seasonal component. In parallel, the trend component is predicted by a linear layer that maps the historical trend to future time steps. The seasonal and trend predictions are then summed to obtain the final forecast $\mathbf{X}_{\text{pred}} \in \mathbb{R}^{O \times D}$.

% The final prediction $\mathbf{X}_{\text{pred}} \in \mathbb{R}^{O \times D}$ is obtained by summing the seasonal and trend forecasts.

\subsection{Extreme Adaptive Attention}

The standard scaled dot-product attention is defined as
\begin{equation}
Q, K, V = \mathrm{Linear}(X^{d}_{enc}),
\end{equation}
\begin{equation}
\mathrm{Attention}(Q,K,V)=\mathrm{Softmax}\left(\frac{QK^{\top}}{\sqrt{d_k}}\right)V,
\end{equation}
where $Q$, $K$, and $V$ denote the queries, keys, and values obtained from the embedded input sequence of the $d$-th series, denoted by $X^{d}_{enc}\in\mathbb{R}^{c\times N_{enc}\times p}$. As in DozerAttention, we flatten the feature-map and patch-size dimensions so that $X^{d}_{enc}\in\mathbb{R}^{N_{enc}\times (c\times p)}$, where each token corresponds to the latent representation of a patch of length $p$. The scaling factor $d_k$ denotes the dimensionality of the query and key vectors.

\begin{figure}[!tb]
    \centering
    \includegraphics[width=\linewidth]{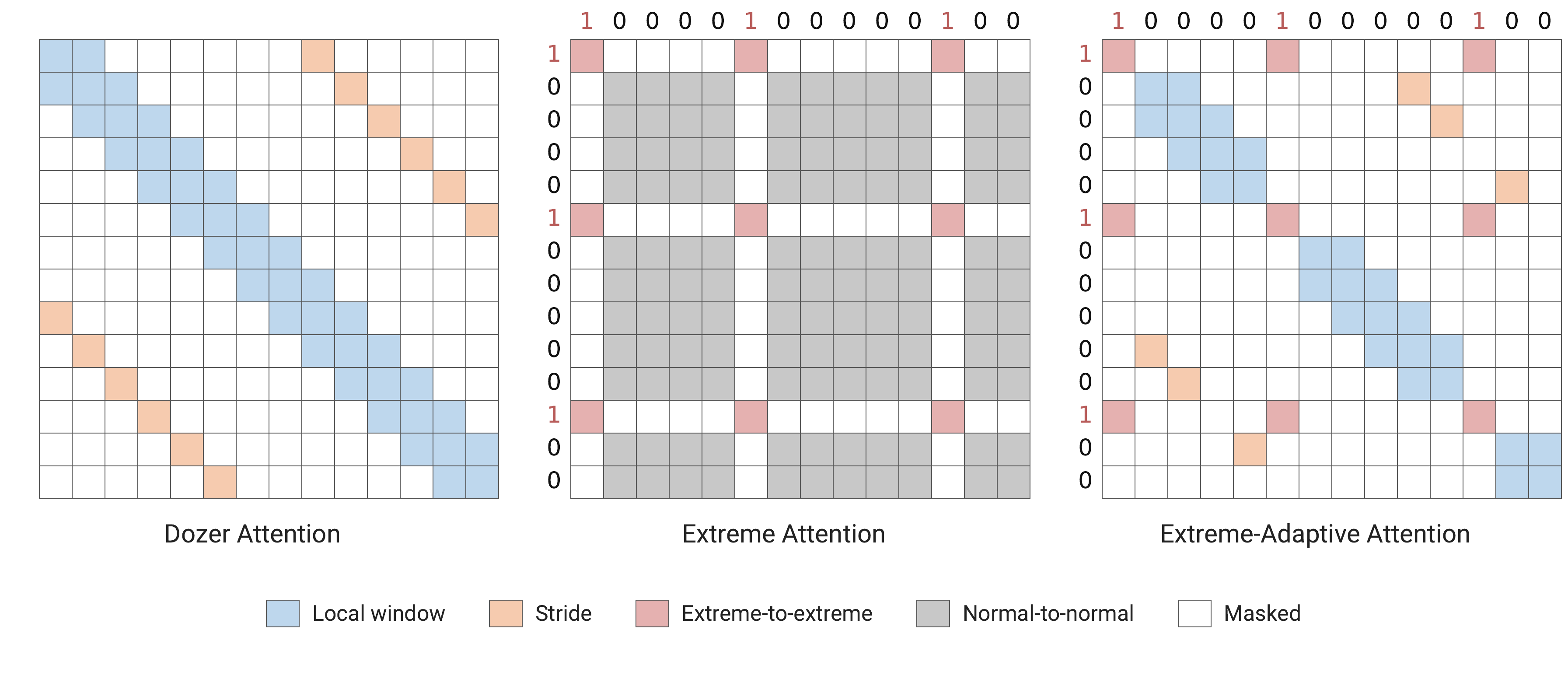}
    \caption{Illustration of attention masks used in Dozer, Extreme, and the proposed Extreme-adaptive Attention.}
    \label{fig:attention_masks}
\end{figure}

Although DozerAttention~\citep{dozer} reduces redundant computation by restricting attention to sparse temporal patterns, it does not explicitly distinguish between normal and extreme input tokens. In highly skewed hydrologic time series data, this limitation is important because extreme observations are rare but often contain more informative patterns than normal observations. To address this issue, we propose Extreme-Adaptive Attention, which consists of three sparse components: Local, Stride, and Extreme, as illustrated in Figure~\ref{fig:attention_masks}. The Local and Stride components are applied only when the query corresponds to a \emph{normal} token, whereas the Extreme component is applied when the query corresponds to an \emph{extreme} token.

% Let $e_{\tau} \in {0,1}$ indicate whether the time step $\tau$ is normal or extreme, where $e_{\tau}=0$ denotes a normal time step and $e_{\tau}=1$ denotes an extreme time step. We define the sets of normal and extreme time-step indices as
% [
% $\mathcal{N} = \{ \tau : e_{\tau}=0 \}$, \qquad
% $\mathcal{E}=\{\tau : e_{\tau}=1\}$.
% ]

\paragraph{Local}
The Local component captures short-range temporal dependencies by allowing a query corresponding to a normal token to attend only to nearby normal keys within a predefined temporal window.

% Let $e_i \in \{0, 1\}$ indicate whether the $i$-th (temporal) patch token is normal or extreme. We define $\mathcal{N} = \{i : e_i = 0\}$ and $\mathcal{E} = \{i : e_i = 1\}$ as the sets of normal and extreme patch token indices, respectively.
% Let $e_p \in \{0,1\}$ indicate whether the temporal patch at index $p$ is normal or extreme. We define $\mathcal{N}=\{p:e_p=0\}$ and $\mathcal{E}=\{p:e_p=1\}$ as the sets of normal and extreme temporal patch indices, respectively. Equation \eqref{eq:local_comp} defines the Local component as follows:

Let $e_i, e_j \in {0,1}$ denote the patch-level label of the temporal patch at index $i$ and $j$, where $0$ indicates that the patch represents a normal event and $1$ indicates an extreme event.  Equation~\eqref{eq:local_comp} defines the Local component as follows:

% We define $\mathcal{N}=\{m \mid e_m=0\}$ and $\mathcal{E}=\{m \mid e_m=1\}$ as the sets of normal and extreme temporal patch indices, respectively.

\begin{equation}
\label{eq:local_comp}
A^{\textit{local}}_{i,j} =
\begin{cases}
q_i \ast k_j, & \text{if } j \in \{|i - j| \le \lfloor w/2 \rfloor\} , \;\; e_i = e_j = 0\\
0, & \text{otherwise}\
\end{cases}
\end{equation}

Where $A$ denotes the attention matrix, whose entries represent the production between queries and keys, and $w$ denotes the local window size. The subscripts $i$ and $j$ denote the temporal index of the query and key vectors, respectively.

\paragraph{Stride}
Time series data often exhibit recurring seasonal patterns. To capture these periodic dependencies, we define the Stride component, in which each normal token query attends only to normal keys at fixed temporal intervals. The Stride component is defined as:
\begin{equation}
A^{\textit{stride}}_{i,j} =
\begin{cases}
q_i \ast k_j, & \text{if } j \in \{|i - j| \bmod s = 0\}, \;\; e_i=e_j=0 \\
0, & \text{otherwise}
\end{cases}
\end{equation}
where $s$ denotes the stride size.

The Dozer mask $M_D$, which is obtained from dozer attention based on the local and the stride components. It allows each query to attend only to nearby keys and keys positioned at fixed temporal intervals, thereby capturing short-range and periodic dependencies while avoiding unnecessary query-key computations. However, $M_D$ treats all time steps equally and does not consider whether a token corresponds to a normal or extreme region.

\paragraph{Extreme-Adaptive}
% Extreme values in time series are known to cluster rather than occur independently~\cite{ferreira2023}. 
To model extreme values in time series, we introduce an \textit{Extreme} component that restricts attention such that extreme queries attend only to other extreme keys:
\begin{equation}
A^{\textit{ext}}_{i,j} =
\begin{cases}
q_i \ast k_j, & \text{if } e_i= e_j \\
0, & \text{otherwise}
\end{cases}
\end{equation}

% The extreme mask $M_E$ represents the extreme-aware label mask $M_E$ defined as follows:
% \begin{equation}
% M_E(i,j)=
% \begin{cases}
%     1,  \text{if } e_i=e_j,\\
%     0,  \text{if } e_i \neq e_j,
% \end{cases}
% \end{equation}
where $e_i, e_j \in {0,1}$ denote the normal/extreme labels of the input tokens at indices $i$ and $j$, respectively. Here, $0$ indicates that the corresponding patch represents a normal event, whereas $1$ indicates an extreme event.
which is constructed from binary labels assigned to each patch token. The extreme-aware mask $M_E$ identifies valid query-key pairs according to these labels, so that normal tokens are separated from extreme tokens during attention computation.

The final mask, denoted as $M_{EA}$, integrates the sparse temporal structure of the Dozer mask with the label constraint imposed by the extreme-aware mask $M_D$. Specifically, the extreme-adaptive attention mask $M_{EA}$ is obtained by applying element-wise operations between the Dozer mask $M_D$ and the extreme-aware mask $M_E$, as follows:

% \begin{equation}
%     M_{\text{EA}} = M_D \land M_E
% \end{equation}

\begin{equation}
    M_{\text{EA}}(i,j)=
\begin{cases}
M_D(i,j)\land M_E(i,j), & e_i=0,\\
M_E(i,j), & e_i=1.
\end{cases}
\end{equation}

for normal queries with $e_i=0$, an element-wise AND operation is applied, such that the query-key product is computed only when the corresponding entries in both $M_D$ and $M_E$ are nonzero. This enables the attention mechanism to capture local and seasonal temporal dependencies among normal events. 
For extreme queries with $e_i=1$, each query attends only to keys with extreme labels ($e_j=1$), allowing the model to capture dependencies among rare extreme events while reducing the influence of the majority normal events. 

As a result, the proposed extreme-adaptive attention mechanism retains the efficient short-range and periodic patterns of DozerAttention while explicitly preserving normal-to-normal and extreme-to-extreme dependencies. Compared with full attention, it avoids computing redundant query-key pairs by preserving only the entries specified by the query-adaptive sparse mask. Compared with the original DozerAttention, it further emphasizes informative extreme-event patterns while suppressing less relevant interactions that may dilute attention to rare extreme tokens.

\section{Experiments}
\paragraph{Datasets}
We conducted experiments on four groups of hydrologic datasets~\citep{pfformer} collected in Santa Clara County, California: Ross, Saratoga, UpperPen, and SFC, each named after its monitoring location. Each group contains a streamflow series and its corresponding rainfall series. The forecasting task focuses on predicting streamflow during the wet season of the hydrologic year, excluding the summer period, specifically from September 2021 to May 2022, using a rolling forecasting setting. The training and validation samples were randomly drawn from data spanning January 1988 to August 2021. We evaluate forecasting performance using Root Mean Squared Error (RMSE) and Mean Absolute Percentage Error (MAPE). Since both streamflow and rainfall are recorded at 15-minute intervals, this corresponds to a long forecasting horizon $(h=288)$.

% Summary statistics of the main streamflow series are reported in Table \ref{tab:stream_stats}

% \begin{table}[t]
% \centering
% \caption{Input Stream Data Statistics}
% \label{tab:stream_stats}
% \renewcommand{\arraystretch}{1.2}
% \begin{tabular}{lcccc}
% \hline
% \textbf{Statistic / Stream} & \textbf{Ross} & \textbf{Saratoga} & \textbf{UpperPen} & \textbf{SFC} \\
% \hline
% min             & 0.00    & 0.00    & 0.00   & 0.00    \\
% max             & 1440.00 & 2210.00 & 830.00 & 7200.00 \\
% mean            & 2.91    & 5.77    & 6.66   & 20.25   \\
% std.\ deviation & 24.43   & 26.66   & 21.28  & 110.03  \\
% skewness        & 19.84   & 19.50   & 13.42  & 18.05   \\
% kurtosis        & 523.16  & 697.78  & 262.18 & 555.18  \\
% \hline
% \end{tabular}
% \end{table}

\paragraph{Baseline Methods}
We evaluate our method against nine strong baselines spanning both general long-term forecasting models and hydrologic prediction methods. FEDFormer~\citep{fedformer} and Informer~\citep{informer} are Transformer-based methods for long-sequence forecasting, with FEDFormer emphasizing frequency-domain decomposition and Informer using ProbSparse self-attention for efficiency. NLinear and DLinear are effective linear forecasting baselines~\citep{linear}, with DLinear explicitly modeling decomposed trend and seasonal components. Attention-LSTM~\citep{Yan2021} is included as a hydrologic multivariate baseline that uses rainfall data to predict streamflow, while NEC+~\citep{nec} is designed specifically for hydrologic series with extreme events. We further include iTransformer~\citep{itransformer}, a recent multivariate forecasting model that inverts the temporal dimension and treats variates as tokens, DAN~\citep{dan}, an extreme-adaptive streamflow forecasting framework based on representation learning and PFformer~\citep{pfformer}, a recent position-free Transformer baseline for hydrologic forecasting.

\paragraph{Implementation Details}
During inference, the model predicts streamflow every 4 hours for the subsequent 3 days. Because both streamflow and rainfall are recorded at 15-minute intervals, this corresponds to a long forecasting horizon $(h=288)$, which falls under the long-term time series forecasting (LSTF) setting. Following the experimental setup mentioned in ~\citep{dan}, all time-series were transformed using the log transform $(x_i = \log(1 + x_i)\,\forall i)$ and then standardized by subtracting the mean and dividing by the standard deviation. To address class imbalance, we incorporated a Kruskal-Wallis sampling strategy~\citep{dan}. We further used a Gaussian Mixture Model (GMM) to derive an outlier score and applied a threshold to obtain labels for normal and extreme events. During inference, predictions were converted back to the original scale by reversing both the standardization and log transformation steps.

\paragraph{Main results}

\begin{table}[!htbp]
\caption{3-day Long-Term ($h = 288$) Series Forecasting Results}
\label{tab:main_results}
\centering
\resizebox{\textwidth}{!}{%
\begin{tabular}{l|c|c|c|c|c|c|c|c}
\toprule
 & \multicolumn{4}{c|}{\textbf{RMSE}} & \multicolumn{4}{c}{\textbf{MAPE}} \\
\cmidrule(lr){2-5} \cmidrule(lr){6-9}
\textbf{Methods} & \textbf{Ross} & \textbf{Saratoga} & \textbf{UpperPen} & \textbf{SFC} & \textbf{Ross} & \textbf{Saratoga} & \textbf{UpperPen} & \textbf{SFC} \\
\midrule
\textbf{FEDformer}    & 6.01 & 6.01 & 3.05 & 23.54 & 2.10 & 1.55 & 1.87 & 2.35 \\
\textbf{Informer}     & 7.84 & 5.04 & 5.88 & 39.89 & 4.05 & 1.43 & 4.10 & 8.64 \\
\textbf{Nlinear}      & 6.10 & 5.23 & 1.57 & 18.47 & 1.99 & 0.83 & 0.45 & 0.92 \\
\textbf{Dlinear}      & 7.16 & 4.33 & 3.53 & 21.62 & 3.10 & 1.40 & 2.35 & 2.74 \\
\textbf{LSTM-Atten}   & 7.35 & 6.49 & 6.35 & 34.17 & 3.74 & 1.80 & 4.76 & 9.90 \\
\textbf{NEC+}         & 9.44 & 1.88 & 2.22 & 17.00 & 4.80 & 0.17 & 0.95 & 1.07 \\
\textbf{iTransformer} & 4.56 & 2.37 & 1.12 & 17.04 & 0.57 & 0.27 & 0.11 & 0.47 \\
\textbf{DAN}          & 4.25 & 1.80 & 1.10 & 15.23 & \underline{0.07} & 0.14 & 0.15 & 0.26 \\
\textbf{PFformer}     &  \underline{4.21} & \underline{1.69} & \underline{1.01} & \textbf{14.98} & 0.10 & \underline{0.10} & \underline{0.06} & \underline{0.18} \\
\midrule
\textbf{Exformer (Ours)}          & \textbf{4.20} & \textbf{1.61} & \textbf{0.96} & \underline{15.12} & \textbf{0.05} & \textbf{0.07} & \textbf{0.04} & \textbf{0.12} \\
\bottomrule
\end{tabular}%
}
\end{table}

Table~\ref{tab:main_results} reports the 3-day forecasting results in terms of RMSE and MAPE, with the best and second-best values for each metric highlighted in bold and underlined, respectively. Overall, Exformer achieves the strongest performance, obtaining the best results in 7 out of 8 reported comparisons. Specifically, Exformer achieves the lowest MAPE on all four datasets and the lowest RMSE on Ross, Saratoga, and UpperPen, while obtaining comparable RMSE performance on SFC. Among the baseline methods, PFformer, DAN, and iTransformer achieve the most competitive results, whereas more general forecasting models, such as FEDformer and Informer, are less effective. These results suggest that 3-day hydrologic forecasting benefits from explicitly modeling rare but important extreme-event patterns.

Compared with PFformer, Exformer improves RMSE by 0.2\%, 4.1\%, and 5.0\% on Ross, Saratoga, and UpperPen, respectively, while reducing MAPE by 50.0\%, 20.0\%, 33.3\%, and 33.3\% on Ross, Saratoga, UpperPen, and SFC, respectively. Compared with DAN, Exformer achieves lower RMSE on all four datasets and reduces MAPE by 28.6\%, 42.9\%, 73.3\%, and 53.8\%, respectively. Exformer also improves over iTransformer, reducing RMSE by 7.9\%, 31.6\%, 14.3\%, and 11.3\% across the four datasets. Overall, these results demonstrate that Exformer provides more accurate 3-day forecasts, particularly when the target series contains rare extreme hydrologic events.

\begin{figure}[!htbp]
\centering
\includegraphics[width=0.75\linewidth]{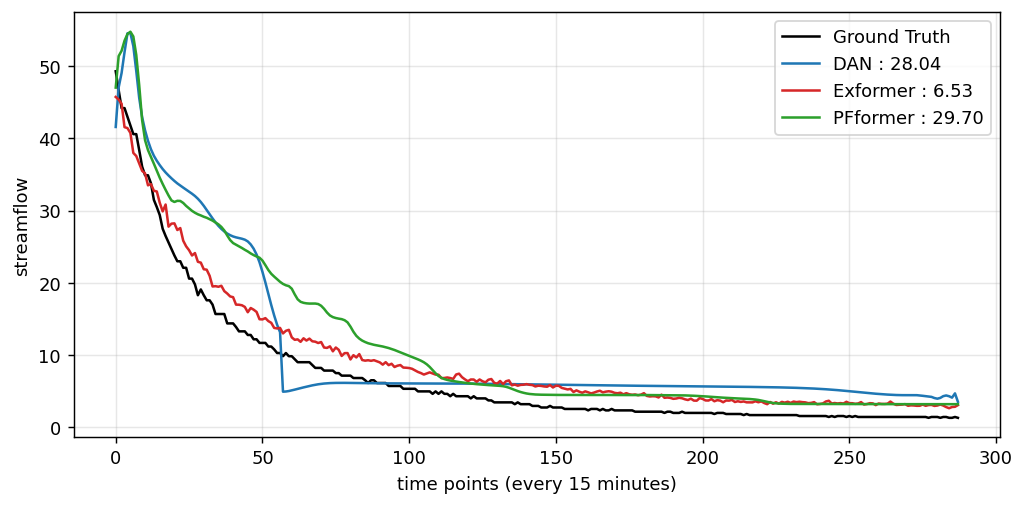}
\caption{
Visualization of forecasting results on the Saragota dataset at a 3-day forecast horizon.
}
\label{fig:case_study}
\end{figure}

Figure~\ref{fig:case_study} visualizes the forecasting results on Saratoga at horizon 288 by comparing the predicted streamflow values from Exformer, DAN, and PFformer against the ground-truth observations. As shown in the figure, Exformer follows the overall temporal pattern more closely and produces predictions that better match the observed streamflow during the forecast period.

\paragraph{Computational efficiency}

We evaluate the computational complexity of Transformer-based methods and present the comparison in Table \ref{tab:complexity}. For self-attention, the proposed Extreme-Adaptive Attention achieves a linear computational complexity with respect to the length of the input sequence $I$. Specifically, the Local and Stride components require at most $\mathcal{O}((w+s)I/p)$, where $w$ and $s$ denote the number of keys attended by each query in the Local and Stride components, respectively. These values are small constants in our experiments, e.g., $w \in \{1,3\}$ and $s \in \{2,3\}$. In contrast, $p$ denotes the patch size and is typically set to a larger value, e.g., $p \in \{24,48,60\}$. Therefore, the coefficient $((w+s)/p)$ remains consistently smaller than 1 in our experimental settings.

In addition, the Extreme component introduces $\mathcal{O}(n_e^2)$ cost by allowing only extreme tokens to attend other extreme tokens, where $n_e$ denotes the number of (temporal) extreme patch tokens. Since extreme events are rare in time-series data, the number of extreme tokens is much smaller than the total number of patch tokens, i.e., $n_e \ll I/p$. Overall, the complexity of the proposed Extreme-Adaptive Attention is $\mathcal{O}((w+s)I/p + n_e^2)$, which substantially reduces computational cost while preserving temporal dependencies among rare extreme patterns.
Here, $(w+s)$ denotes the maximum number of keys attended by each normal query from the Local and Stride components. A normal query attends to exactly $(w+s)$ keys when none of its selected Local and Stride keys corresponds to extreme tokens. However, if one or more selected keys are extreme tokens, those keys are removed, and the query attends to fewer than $(w+s)$ keys. Extreme queries are handled separately by the Extreme component, where extreme tokens attend to one another.

\begin{table}[!htbp]
    \centering
    \caption{Computational complexity of self-attention. $I$ denotes the encoder's input sequence length; $N$ denotes the number of variates.}
    \label{tab:complexity}
    \renewcommand{\arraystretch}{1.2}
    \resizebox{\columnwidth}{!}{%
    \begin{tabular}{lcccccc}
        \toprule
         & \textbf{Transformer} & \textbf{Informer} & \textbf{FEDformer} & \textbf{PFformer} & \textbf{iTransformer} & \textbf{Exformer (Ours)} \\
        \midrule
        \textbf{Self-attention} & $\mathcal{O}(I^2)$ & $\mathcal{O}(I \log I)$ & $\mathcal{O}(I)$ & $\mathcal{O}(I^2)$ & $\mathcal{O}(N^2)$ & $\mathcal{O}((w+s)I/ p + n_e^2)$ \\
        \bottomrule
    \end{tabular}%
    }
\end{table}

% $\mathcal{E} \subseteq \{1, 2, \ldots, I\}$

% We also evaluated the Params (total number of learnable parameters), FLOPs (floating-point operations), and Memory (maximum GPU memory consumption). Table~\ref{tab:model_complexity} presents the quantitative results of this efficiency comparison with a historical length set at 1440 and a forecasting horizon of 288. NLinear and DLinear, characterized by simple architectures that employ a single linear layer to directly generate forecasts from historical records, achieved the best efficiency among all methods considered, with 0.03 GFLOPs, 1.26M parameters, and 15.68 MB of memory. iTransformer also demonstrated competitive efficiency, requiring only 0.34 GFLOPs and 37.80 MB of memory. Notably, the proposed Exformer showed superior efficiency compared with other transformer and LSTM-based baselines. Specifically, Exformer requires only 31.60 GFLOPs, which is approximately 5.9 times less than PFformer, 8.3 times less than DAN, 9.9 times less than Informer, 8.7 times less than FEDformer, and 18.4 times less than NEC+. We observe similar result in memory consumption: Exformer consumes only 240.85 MB, representing a 2.9 times reduction compared to PFformer, a 4.5 times reduction compared to Informer, and a 17.1 times reduction compared to NEC+.

\begin{table}[htbp]
    \centering
    \caption{The Model complexity results for input length of 1440 and forecasting horizon 288.}
    \label{tab:model_complexity}
        \renewcommand{\arraystretch}{1.2}

    \begin{tabular}{lrrr}
        \toprule
        \textbf{Model} & \textbf{FLOPs (G)} & \textbf{Params (M)} & \textbf{Memory (MB)} \\
        \midrule
        Exformer    & 31.60   & 14.30  & 240.85  \\
        PFformer     & 187.07  & 9.21   & 705.34  \\
        DAN          & 261.83  & 31.88  & 3069.17 \\
        Informer     & 313.99  & 7.57   & 1081.30 \\
        FEDformer    & 275.13  & 16.83  & 1001.10 \\
        % NLinear      & 0.03    & 1.26   & 15.68   \\
        DLinear      & 0.03    & 1.26   & 15.68   \\
        NEC+         & 580.97  & 30.13  & 4110.44 \\
        iTransformer & 0.34      &7.19     & 37.80     \\
        \bottomrule
    \end{tabular}
\end{table}

We also evaluated the Params (total number of learnable parameters), FLOPs (floating-point operations), and Memory (maximum GPU memory consumption). Table~\ref{tab:model_complexity} presents the quantitative results of this efficiency comparison with a historical length of 1440 and a forecast horizon of 288. NLinear and DLinear, which use simple linear architectures to directly generate forecasts from historical records, achieved the best efficiency among all methods, with 0.03 GFLOPs, 1.26M parameters, and 15.68 MB of memory. iTransformer also demonstrated competitive efficiency, requiring only 0.34 GFLOPs and 37.80 MB of memory. This is mainly because iTransformer applies self-attention across variates rather than temporal tokens, leading to a complexity of $\mathcal{O}(N^2)$, where $N$ is the number of variates. In our experimental setting, $N$ is only 2, whereas the input sequence length $I$ is 1440. As a result, iTransformer incurs a considerably lower attention cost than temporal attention-based methods, whose complexity scales with the input sequence length $I$.

Although linear models and variate-wise attention methods achieve lower computational cost, their forecasting accuracy is consistently lower than that of Exformer. As shown in Table~\ref{tab:main_results}, Exformer achieves the best RMSE on three of the four datasets and the best MAPE across all four datasets. This indicates that Exformer provides a stronger accuracy-efficiency trade-off: it substantially improves forecasting performance while still maintaining lower computational cost than temporal attention-based Transformer baselines.

Notably, Exformer demonstrated superior efficiency compared with other Transformer and LSTM-based baselines. Specifically, Exformer requires only 31.60 GFLOPs, which is approximately 5.9 times less than PFformer, 8.3 times less than DAN, 9.9 times less than Informer, 8.7 times less than FEDformer, and 18.4 times less than NEC+. Similar results are observed in memory consumption: Exformer consumes only 240.85 MB, representing a 2.9 times reduction compared to PFformer, a 4.5 times reduction compared to Informer, and a 17.1 times reduction compared to NEC+.

% \begin{table}[htbp]
%     \centering
%     \caption{Computational complexity of self-attention. The Encoder's input historical sequence length is denoted $I$; $N$ denotes the number of variates.}
%     \label{tab:complexity}
%     \renewcommand{\arraystretch}{1.2}
%     \begin{tabular}{ll}
%         \toprule
%         \textbf{Method} & \textbf{Self-attention} \\
%         \midrule
%         Transformer     & $\mathcal{O}(I^2)$       \\
%         Informer        & $\mathcal{O}(I \log I)$  \\
%         FEDformer       & $\mathcal{O}(I)$         \\
%         PFformer        & $\mathcal{O}(?)$         \\
%         iTransformer    & $\mathcal{O}(N^2)$         \\
%         Exformer (Ours) & $\mathcal{O}(?)$        \\
%         \bottomrule
%     \end{tabular}
% \end{table}

\subsection{Ablation}
\paragraph{Local, stride, and extreme components ablation}

To evaluate the relative importance of the Local, Stride, and Extreme components, we conducted experiments using only one of these components in extreme adaptive attention. The results are summarized in Table \ref{tab:ablation}. The hyperparameters of each component follow the settings used in the main results. Overall, each component, working alone, performed slightly worse than the full Exformer, demonstrating the effectiveness of each component in capturing essential attributes of the streamflow data. Specifically, the Exformer achieved the best RMSE and MAPE on Ross, Saratoga, and SFC. Among the individual components, the Extreme branch consistently produced the closest performance to the full model. These results confirm that combining the three mechanisms within the Extreme Adaptive Attention gives the best overall performance.
\begin{table}[h]
\centering
\caption{Ablation study of three key components of Extreme Adaptive attention: Local, Stride, and Extreme. Significant values are in bold.}
\label{tab:ablation}
\setlength{\tabcolsep}{7pt}        % horizontal spacing
\renewcommand{\arraystretch}{1.2}   % vertical spacing
\begin{tabular}{lcccccccc}
\toprule
Methods & \multicolumn{2}{c}{Exformer} & \multicolumn{2}{c}{Local} & \multicolumn{2}{c}{Stride} & \multicolumn{2}{c}{Extreme} \\
\midrule
Metric       & {\scriptsize RMSE}            & {\scriptsize MAPE}            & {\scriptsize RMSE}            & {\scriptsize MAPE}            & {\scriptsize RMSE}   & {\scriptsize MAPE}   & {\scriptsize RMSE}            & {\scriptsize MAPE}            \\
\midrule
Ross    & \textbf{4.2}           &  \textbf{0.05}         & 4.2         & 0.08          & 4.21 & 0.07 & 4.2 & 0.06          \\
Saratoga     & \textbf{1.61}          & \textbf{0.07}           & 1.74 & 0.1 & 1.7
 & 0.09 & 1.62          & 0.08          \\
UpperPen   & \textbf{0.96}  & 0.04 &0.98        & 0.04          & 1.02 & 0.05 & \textbf{0.96}          & \textbf{0.03}          \\
SFC     & \textbf{15.12} & \textbf{0.12} & 16.53          & 0.19          &16.07 &0.13 & 15.21
         & 0.11         \\
\bottomrule
\end{tabular}
\end{table}

% \begin{table}[t]
% \centering
% \caption{Comparison of \textit{extreme adaptive} Dozerformer with other attention mechanisms for prediction length 288. Significant values are in bold.}
% \label{tab:attn_comp}
% \setlength{\tabcolsep}{5pt}
% \renewcommand{\arraystretch}{1.15}
% \small
% \begin{tabular}{l cc cc cc cc}
% \toprule
% Methods
% & \multicolumn{2}{c}{Ext-Adapt}
% & \multicolumn{2}{c}{Dozer}
% & \multicolumn{2}{c}{Extreme}
% & \multicolumn{2}{c}{Canonical} \\
% \cmidrule(lr){1-1} \cmidrule(lr){2-3} \cmidrule(lr){4-5} \cmidrule(lr){6-7} \cmidrule(lr){8-9}
% Metric
% & RMSE & MAPE
% & RMSE & MAPE
% & RMSE & MAPE
% & RMSE & MAPE \\
% \midrule
% Ross      & 4.2 & \textbf{0.05} & 4.2 & 0.08 & 4.2
%  & \underline{0.06} & 4.2 & 0.08 \\
% Saratoga  & \textbf{1.62} & \textbf{0.08}  & 1.7 &  0.09 & \underline{1.65} & \textbf{0.08} & 1.7 &  0.09  \\
% UpperPen  & \textbf{0.96} & 0.04 & \underline{0.97} & 0.04 & 0.99 & 0.04 & 0.99 & 0.04 \\
% SFC       & \textbf{15.12}  & \textbf{0.12} & \underline{15.28} & \underline{0.13} & 16.02 & 0.16 & 15.33 & 0.13 \\
% \bottomrule
% \end{tabular}
% \end{table}
\begin{table}[t]
\centering
\caption{Comparison of \textit{extreme adaptive} Exformer with other attention mechanisms for prediction length 288. Significant values are in bold.}
\label{tab:attn_comp}
\renewcommand{\arraystretch}{1.2}
\setlength{\tabcolsep}{5pt}
\footnotesize
\begin{tabular}{l cc cc cc cc cc cc}
\toprule
Methods
& \multicolumn{2}{c}{Ext-Adapt}
& \multicolumn{2}{c}{Dozer}
& \multicolumn{2}{c}{Canonical}
& \multicolumn{2}{c}{AutoCorr}
& \multicolumn{2}{c}{FedAttn}
& \multicolumn{2}{c}{ProbSparse} \\
\cmidrule(lr){1-1} \cmidrule(lr){2-3} \cmidrule(lr){4-5} \cmidrule(lr){6-7} \cmidrule(lr){8-9} \cmidrule(lr){10-11} \cmidrule(lr){12-13}
Metric
& {\scriptsize RMSE} & {\scriptsize MAPE}
& {\scriptsize RMSE} & {\scriptsize MAPE}
& {\scriptsize RMSE} & {\scriptsize MAPE}
& {\scriptsize RMSE} & {\scriptsize MAPE}
& {\scriptsize RMSE} & {\scriptsize MAPE}
& {\scriptsize RMSE} & {\scriptsize MAPE} \\
\midrule
Ross      & 4.2 & \textbf{0.05} & 4.2 & 0.08 & 4.21 & 0.08 & 4.2 & 0.08
&4.2 & 0.07& 4.2 & 0.07\\
Saratoga  & \textbf{1.61} & \textbf{0.07} & 1.7 & 0.09 & 1.7 & 0.09 &1.71 & 0.09 &1.71 & 0.09 &1.7  &0.09 \\
UpperPen  & \textbf{0.96} & 0.04 & 0.97& 0.04 & 0.99 & 0.04 & 0.98 & 0.04 & 0.98 & 0.04 & 0.98 & 0.04\\
SFC       & \textbf{15.12} & \textbf{0.12} & 15.28 & 0.13 & 15.33 & 0.13 & 15.48 & 0.13& 15.51& 0.13&  15.26& 0.13\\
\bottomrule
\end{tabular}
\end{table}

\paragraph{Attention mechanism comparison} To demonstrate the effectiveness of the proposed Extreme adaptive attention mechanism, we conducted a comparative analysis by replacing it with other attention mechanisms within the framework. Specifically, we compared it with the standard Dozer attention~\citep{dozer}, canonical full attention~\citep{transformer}, Auto-Correlation~\citep{autoformer}, Frequency Enhanced Attention (FedAttn)~\citep{fedformer}, and ProbSparse attention~\citep{informer}. The results are presented in Table~\ref{tab:attn_comp} for a prediction length of 288.

Extreme-Adaptive Attention consistently outperformed all other mechanisms, achieving the lowest RMSE and MAPE in three of the four datasets (Saratoga, UpperPen, and SFC). In particular, on the Saratoga dataset, the extreme adaptive variant reduced RMSE by 4.7\% compared to standard Dozer attention. On the SFC dataset, it achieved a 1.0\% improvement over standard Dozer and a 2.4\% improvement over Auto-Correlation. For the Ross dataset, all methods achieved identical RMSE of 4.2, yet the extreme adaptive attention achieved a notably lower MAPE of 0.05, representing a 37.5\% reduction compared to the standard Dozer and canonical attention (both 0.08). Auto-Correlation and FedAttn performed comparably across all datasets but consistently lagged behind the extreme adaptive variant. ProbSparse attention achieved comparable performance on SFC but remained slightly below that of the proposed method. Overall, these results show that the proposed attention mechanism achieves a better balance between capturing general temporal dependencies and informative extreme patterns, leading to better forecast performance, particularly in the presence of extreme events.

\subsection{Parameter sensitivity}
\begin{figure}[htbp]
    \centering
    \includegraphics[width=0.32\textwidth]{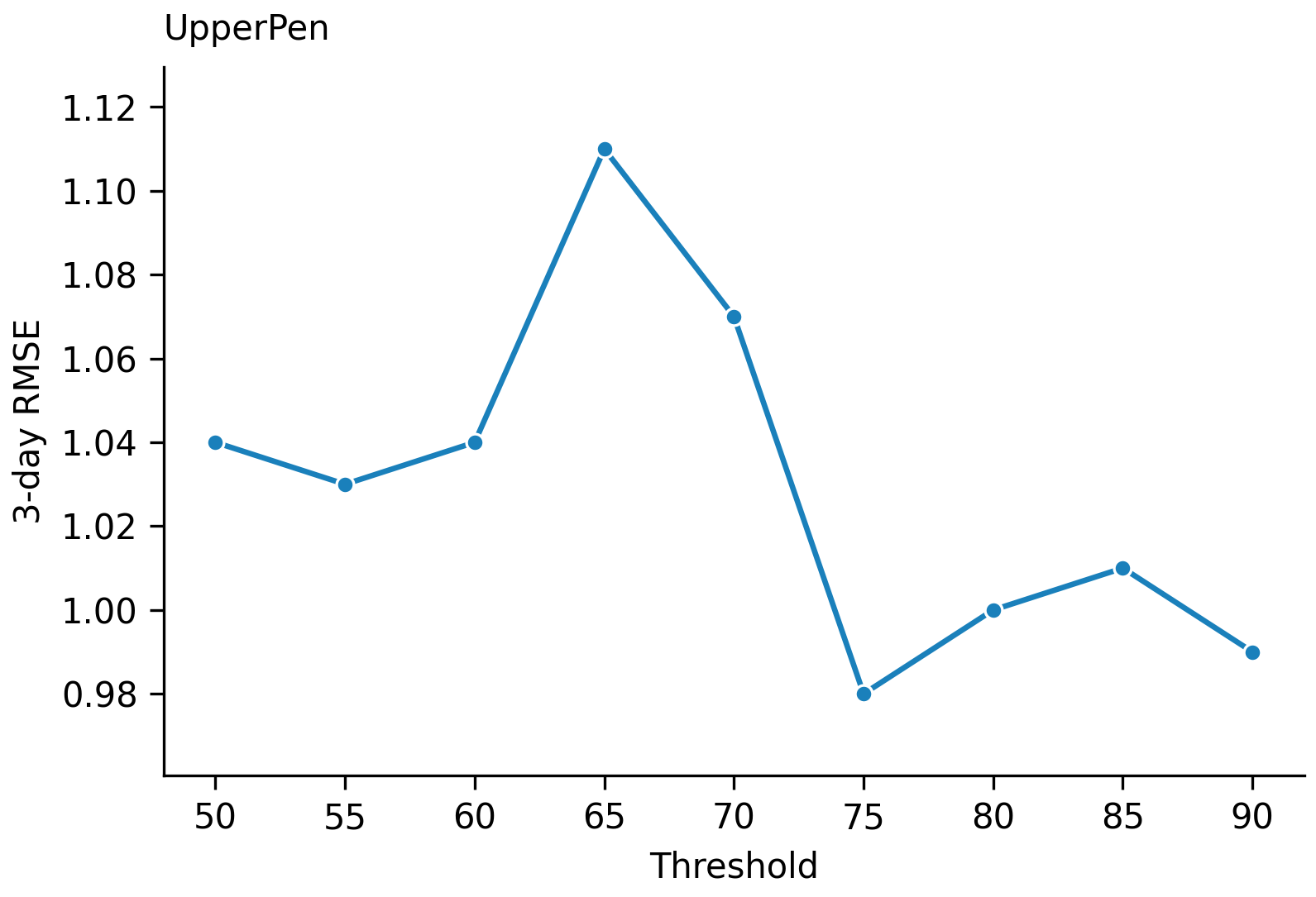}
    \hfill
    \includegraphics[width=0.32\textwidth]{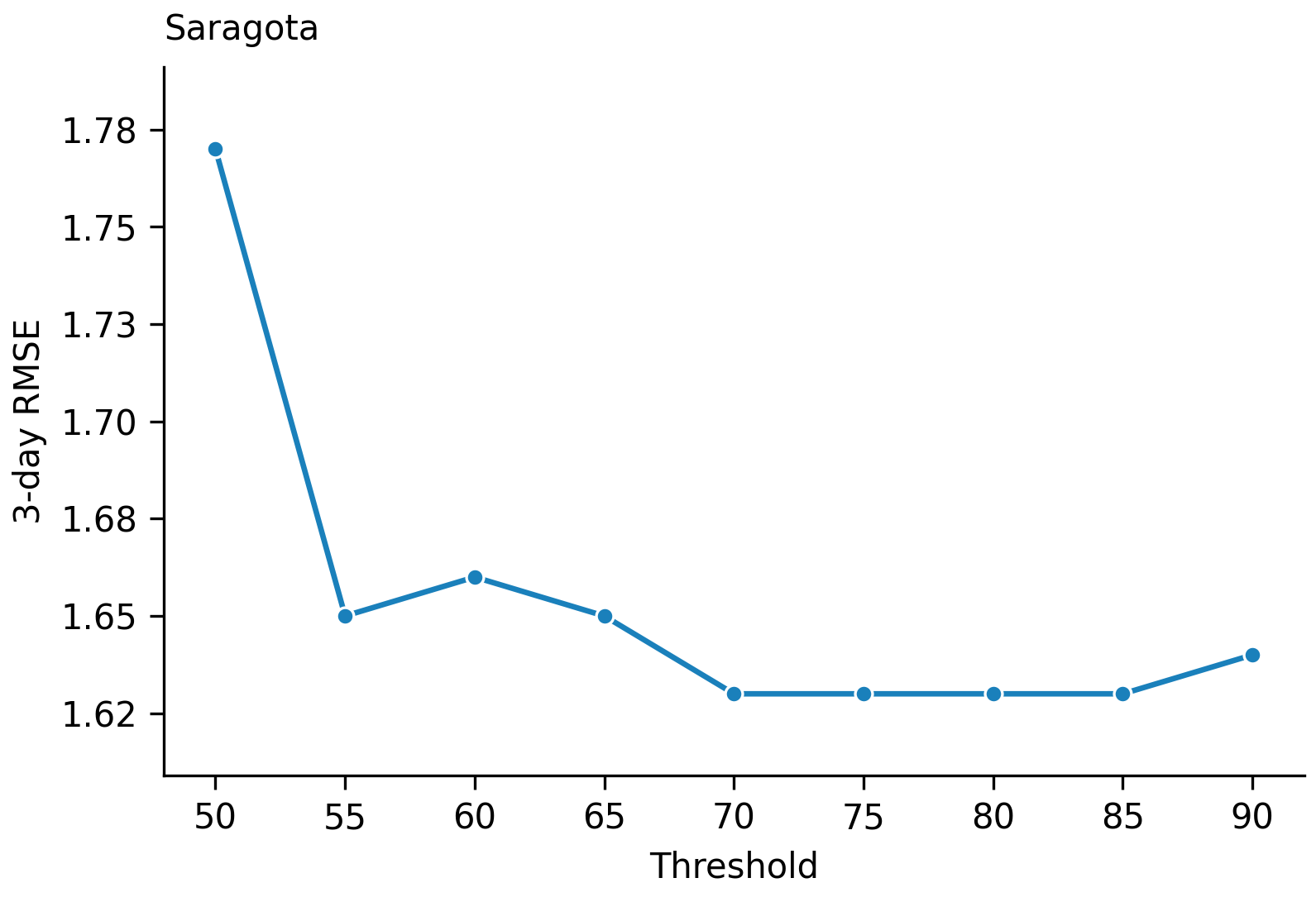}
    \hfill
    \includegraphics[width=0.32\textwidth]{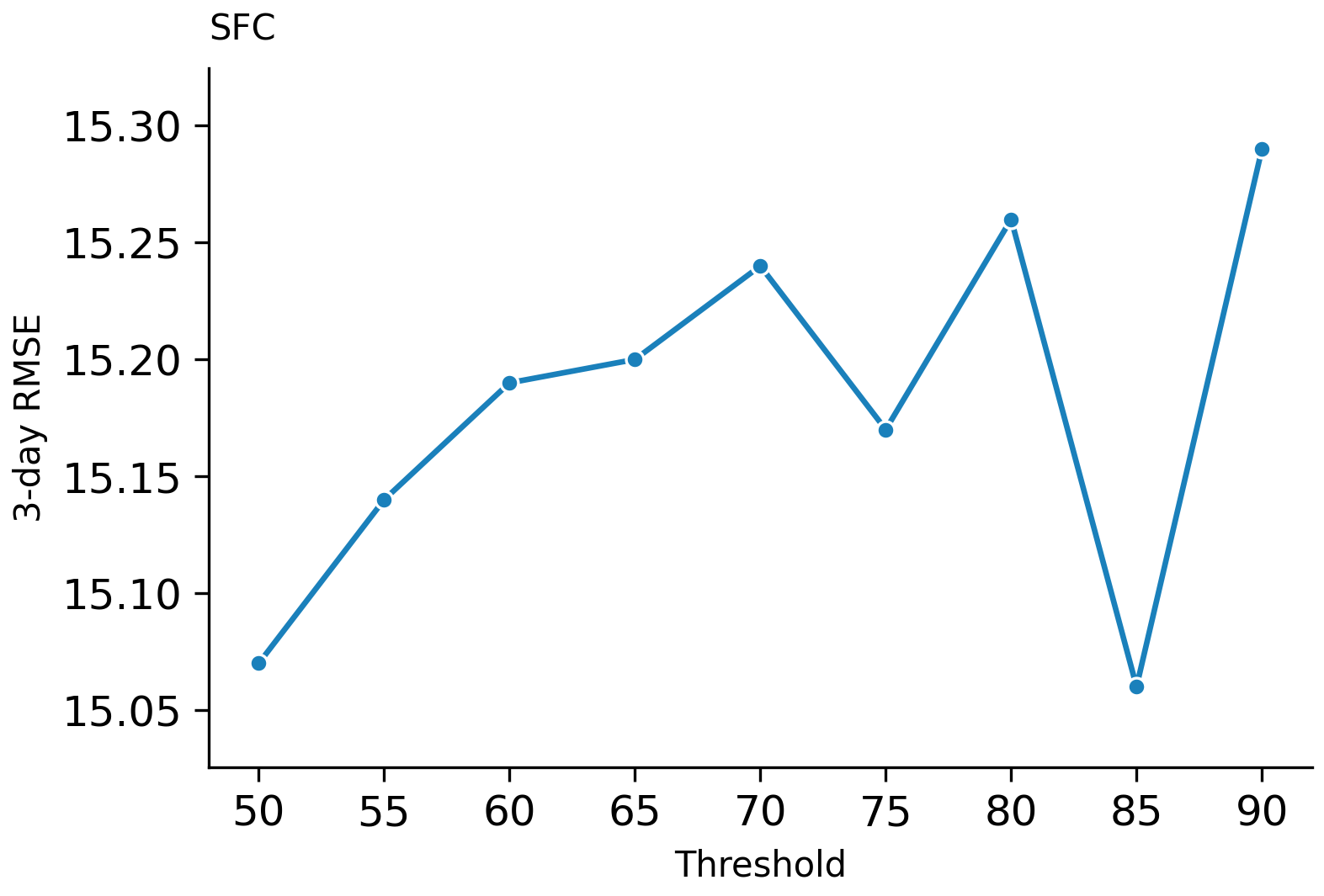}
    \caption{Sensitivity analysis of 3-day prediction across different threshold values}
    \label{fig:param_sensitivity}
\end{figure}

\paragraph{Effect of extreme event threshold on the accuracy} 
We investigate the sensitivity of our model to the choice of threshold percentile $k$ applied to the outlier scores derived from the GMM model for labeling extreme events. All values greater than the $k^{\text{th}}$ percentile are marked extreme. We vary $k \in \{50, 55, 60, \ldots, 90\}$ and report the resulting 3-day RMSE as shown in Figure \ref{fig:param_sensitivity}. The results show that forecasting accuracy is influenced by the choice of threshold across the tested range. Specifically, we observe two distinct patterns: when the threshold is relatively low, a larger fraction of timesteps is labeled as extreme, which makes the extreme-event label less informative and leads to higher errors. Higher thresholds, which label only the most anomalous points, generally yield lower errors and the best overall performance. We also observe that at very high thresholds, the error may rise slightly on some datasets, as the extreme-event class becomes too low to provide a sufficient learning signal. This shows that the model works better when extreme events are defined more selectively, so the label captures accurate anomalous behavior rather than typical variations, while remaining effective across a wide range of threshold values.

\section{Conclusion}

In this paper, we proposed Exformer, an Extreme-Adaptive Transformer for multi-step forecasting on highly skewed hydrologic time series. Unlike standard Transformer models that apply the same attention pattern to all tokens, Exformer introduces an extreme-adaptive attention mechanism that distinguishes between normal and extreme patch tokens. The proposed mechanism integrates Local, Stride, and Extreme components, allowing normal query tokens to capture short-range and periodic dependencies while enabling extreme query tokens to selectively attend to other extreme tokens. Compared with state-of-the-art baselines, including DAN and PFformer, Exformer achieves the best results in most RMSE and MAPE comparisons, demonstrating its effectiveness in modeling rare extreme hydrologic events. In addition to improving forecasting accuracy, Exformer shows strong computational efficiency, requiring substantially fewer FLOPs and less memory than several Transformer- and LSTM-based extreme-adaptive baselines. These results suggest that query-adaptive extreme-aware attention is an effective direction for forecasting imbalanced and highly skewed time series.

% Future work may extend Exformer to investigate more adaptive extreme-token identification strategies and explore dynamic attention patterns that can further improve robustness under varying temporal conditions.

% \section*{Acknowledgments}
% We are thankful to Howard Yen, Howard Chen, Lucy He, Mengzhou Xia, Tianyu Gao, Xi Ye, Yihe Dong, and the members of the Princeton NLP group for their helpful comments and discussion. %
% DF is partly supported by a Google PhD Fellowship.
% This research is also funded by the National Science Foundation (IIS-2211779).

\bibliography{ref}
\bibliographystyle{iclr2026_conference}

% \newpage
% \input{sections/10-appendix}

\end{document}